# Neural Network Architecture Search Enabled Wide-Deep Learning (NAS-WD) for Spatially Heterogenous Property Awared Chicken Woody Breast Classification and Hardness Regression

Chaitanya Pallerla[a,b], Yihong Feng[a], Casey M. Owens[c], Ramesh Bahadur Bist[a], Siavash Mahmoudi[a], Pouya Sohrabipour[a], Amirreza Davar[a], Dongyi Wang[a,b,*]

[a] Department of Biological and Agricultural Engineering, University of Arkansas, Fayetteville, AR, USA, 72701

[b] Department of Food Science, University of Arkansas, Fayetteville, AR, USA, 72701

[c] Department of Poultry Science, University of Arkansas, Fayetteville, AR, USA, 72701

*Corresponding: dongyiw@uark.edu*

**ABSTRACT:** Due to intensive genetic selection for rapid growth rates and high broiler yields in recent years, the global poultry industry has faced a challenging problem in the form of woody breast (WB) conditions. This condition has caused significant economic losses as high as $200 million annually, and the root cause of WB has yet to be identified. Human palpation is the most common method of distinguishing a WB from others. However, this method is time-consuming and subjective. Hyperspectral imaging (HSI) combined with machine learning algorithms can evaluate the WB conditions of fillets in a non-invasive, objective, and high-throughput manner. In this study, 250 raw chicken breast fillet samples (normal, mild, severe) were taken, and spatially heterogeneous hardness distribution was first considered when designing HSI processing models. The study not only classified the WB levels from HSI but also built a regression model to correlate the spectral information with sample hardness data. To achieve a satisfactory classification and

abstractregression model, a neural network architecture search (NAS) enabled a wide-deep neural network model named NAS-WD, which was developed. In NAS-WD, NAS was first used to automatically optimize the network architecture and hyperparameters. The classification results show that NAS-WD can classify the three WB levels with an overall accuracy of 95%, outperforming the traditional machine learning model, and the regression correlation between the spectral data and hardness was 0.75, which performs significantly better than traditional regression models.

# 1. Introduction

Poultry is the most popular proteinaceous food consumed by most people in the United States and worldwide. In 2021, according to the United States Department of Agriculture (USDA, 2022), the availability of chicken per American was 68.1 pounds (on a boneless, edible basis). From 1910 to 2021, the per capita of chicken exponentially increased compared to the per capita of beef and pork, which shows the higher demand for chicken. To meet the increasing demands, geneticists have been forced to improve growth and breast yields (Petracci et al., 2019). These rapidly growing broilers have greater breast yield with increased pectoralis muscles that were prone to be affected by myopathy disorders. Many believe that myopathy is a muscle disease in chickens caused by the frequent contraction of muscles, which turns into woody breast myopathies (Caldas-Cueva and Owens, 2020; Kang et al., 2020). There are many types of myopathies, among which woody breast (WB) is the most common. Woody breast condition is undesirable and indicates the inferior quality of the chicken breast (Caldas-Cueva and Owens, 2020). Some characteristics of WB are increased free water, which causes increased drip loss and water loss upon cooking (Yang et al., 2021). In 2016, incidences as high as 30 to 50% of severe woody breast conditions were reported by the U.S broiler industry, and this was seen in the broilers that were grown for 8 weeks to a live weight of >4.2 kg prior to slaughter (Tijare et al., 2016). This adds to processing costs in the broiler industry as fillet samples need to be identified using a subjective labor-intensive method to assess hardness for quality evaluation purposes (Sun et al., 2018). It is estimated that this WB condition has caused significant economic losses as high as $200 million for the national poultry industry (Kuttappan et al., 2016) and over $1 billion in direct and indirect costs across the global poultry industry (Barbut, 2020).

Many methods of detecting and classifying WB myopathy include subjective methods, objective instrumentation-based invasive measurements, and rapid, non-invasive in-line methods (Caldas-Cueva and Owens, 2020). Firstly, subjective methods include visual and tactile methods, which are still the mainstream methods used in poultry industries. As visual identification and hand palpation need some training and expertise to detect and classify WB myopathy, there is a chance of a high misclassification rate. In addition, these subjective methods are also very labor intensive, which adds to the cost of the product. Based on the tactile evaluation, the woody breast can be categorized using a 4-point scale (Normal = 0, Mild = 1, Moderate = 2, and Severe = 3) (Tijare et al., 2016). Secondly, invasive methods use instruments such as texture analyzers (Mudalal et al., 2015) and pH meters to detect the difference between the normal and WB based on characteristics like hardness, pH, color, and other biomarkers like protein content and water holding capacity (WHC). Chicken breast affected by woody breast condition shows significant color changes with increased $L^*$, $a^*$, $b^*$, and pH values (Geronimo et al., 2019), and the WB has hardness and chewiness higher than the normal chicken breast (Chatterjee et al., 2016). Although these results are promising, these instrumentation-based methods are very time-consuming in sample preparation and analysis for texture, pH, and color measurements. Bioelectrical impedance analysis (BIA) (Seafood Analytics, Version 3.0.0.3, Clinton Town, MI, USA) combined with supervised and unsupervised machine learning algorithms were reported to be effective in identifying the woody breast. Four electrodes were inserted into the geometric center of each breast fillet along the ventral surface, and the device measured the corresponding resistance data (Siddique et al., 2022). Still, BIA measurements can be influenced by bird size, hydration status, and environmental conditions. (Walter et al., 2011, Catapano, et al., 2023)

To solve the challenges of the subjective and destructive methods mentioned above, there is an urgent need to develop rapid detection methods such as computer vision systems (CVS). The CVS reported include near-infrared (NIR) spectroscopy (Geronimo et al., 2019), Optical coherence tomography (OCT) (Yoon et al., 2016), and Hyperspectral imaging (HSI). Woody breast conditions in 8-week-old broiler carcasses were observed with conformational changes, which can be identified using image analysis (Caldas-Cueva and Owens, 2020). In addition, fillets have different bending properties, which can be determined by side-view imaging technology (Yoon et al., 2022). NIR spectroscopy was very good at detecting WB myopathy by correlating the spectral information with biomarkers such as protein content and water-holding capacity (WHC). Later, it was found that this method was only accounting for the biomarkers' 1 cm depth (about 0.39 in) from the surface of the breast fillets while these biomarkers change with depth and spatial locations on the breast, which made NIR methods unreliable where the compositions are heterogeneous (Wold et al., 2019). When it comes to spectral-domain OCT imaging systems it is a real-time imaging technique that can be used to detect early-stage woody breast myopathy (WBM) lesions and types of WBM lesions (Yoon et al., 2016). In contrast, the major disadvantage is that OCT misclassifies the normal and WB if they have excessive adipose and connective tissue (Yoon et al., 2016) and the complex surface topography of meat tissue (Yoon et al., 2016). Whereas integrating image analysis algorithm as a part of the OCT system (Doc L-Pix model, Loccus Biotecnologia, Sao Paulo, Brazil) has shown a classification accuracy of 91.8% (Geronimo et al., 2019) but it was found that the classification results were based on image quality and illumination where the price range of basic OCT system $20,000 to $50,000. Therefore, hyperspectral imaging system has different benefits over these imaging technologies and is widely used.

Hyperspectral imaging (HSI) is a rapid and non-destructive method that simultaneously captures an object's spectral and spatial data, allowing for the identification and detection of many entities. HSI stores information about light intensity as a function of wavelength in the form of a hypercube consisting of spectral (with a resolution of 1-10 nm) and spatial data. HSI has advantages over other non-destructive technologies, such as minimal requirements of sample preparation. The rich information contained in the spectral data can benefit process monitoring, real-time inspection, qualitative and quantitative assessments, and chemical mapping capabilities. Because of these benefits, HSI can be used to classify food products according to food quality, consumer preferences, and other product requirements (Elmasry et al., 2012). However, a previous study showed that there was no clear difference between the mean hyperspectral imaging information of normal and WB samples (Yoon et al., 2016), which could be caused by the spatially heterogeneous properties and the improper calibrations, both may significantly affect the representativeness of the mean spectral information. In addition, the selection of hyperspectral imaging processing algorithms can also impact the interpretation and analysis of the spectral data.

This study considers the spatially heterogeneous properties of fillet samples when conducting WB modeling using HSI. In addition, besides directly classifying the woody breast levels from the hyperspectral imaging system, the hyperspectral signals were correlated with hardness data based on regression models, and raw fillet hardness distribution maps were generated. The heterogeneous distribution of hardness data is considered the spatial outcomes of chemical component variations. In the study, we considered the fillet spatial information and developed a novel neural network architecture search (NAS) enabled wide-deep neural network (NAS-WD) model for correlating the spectral data and hardness to create a hardness distribution map and to classify the woody breast levels. Compared to the widely used multi-layer perception-based neural network model, the NAS-WD model has two main improvements, including the WD module and the NAS model. The WD module considers both the linear and non-linear relationships between the input features and output hardness. The model uses joint training, which optimizes all the parameters accounting for both the wide and deep parts simultaneously during training, making it perfect for correlating the complex relations between the data. The concept of WD learning models has been increasingly applied in various scientific fields to address complex prediction and classification tasks. Wilson et al. (2021) developed a WD Learning model for automatic cell type identification, achieving superior classification accuracy and increasing overall cell type prediction accuracy from 36.5% to 86.9% compared to existing models. WD learning even outperformed deep learning models, CHETAH (70.36% accuracy), and SingleR (70.59% accuracy) in discriminating between similar cell types and handling cross-platform differences in single-cell RNA sequencing technologies. Alatrany et al. (2023) developed an Alzheimer's disease prediction model using a hybrid feature selection approach with WD neural networks, achieving 99% accuracy and F1-score. The model outperformed Random Forest (89% accuracy, 88% F1-score), Wide Neural Network (94% accuracy, 93% F1-score), and Deep Neural Network (93% accuracy, 92% F1-score). Cheng et al. (2016) developed a WD Learning model to improve the app acquisition rate and serving latency compared to wide-only and deep-only models. Key findings include that the WD model effectively balances memorization and generalization, optimizes performance through joint training, and efficiently handles large-scale data with over 500 billion examples. Wide and Deep Learning outperformed Generalized Linear Models (GLMs), Deep Neural Networks (DNNs), and Factorization Machines, demonstrating superior recommendation performance and scalability.

In this study, we first integrated the neural network architecture search (NAS) algorithm into wide and deep learning to optimize the network architecture design and build a new model named NAS-WD, whose overall diagram is shown in Fig. 1. The first objective of this study is to develop NAS-WD model to identify and classify the fillet's woody breast condition into three manual graded categories (Normal breast (NB), Mild Woody breast (MWB), and Severe Woody breast (SWB)) based on the HIS signals. The second objective is to develop a NAS-WD-based regression model to correlate the spectral signals with the cranial region's compression force measured from texture analyzers, considering the cranial region is the common region for all the severity levels where the hardness changes significantly (Tijare et al., 2016). A hardness distribution map could be further generated by applying the regression model to the entire HSI cubic. To prove the effectiveness of NAS-WD, the performance of classic supervised machine learning algorithms, including support vector machine (SVM), multi-layer perceptron (MLP), and partial least square regression (PLSR), were compared.

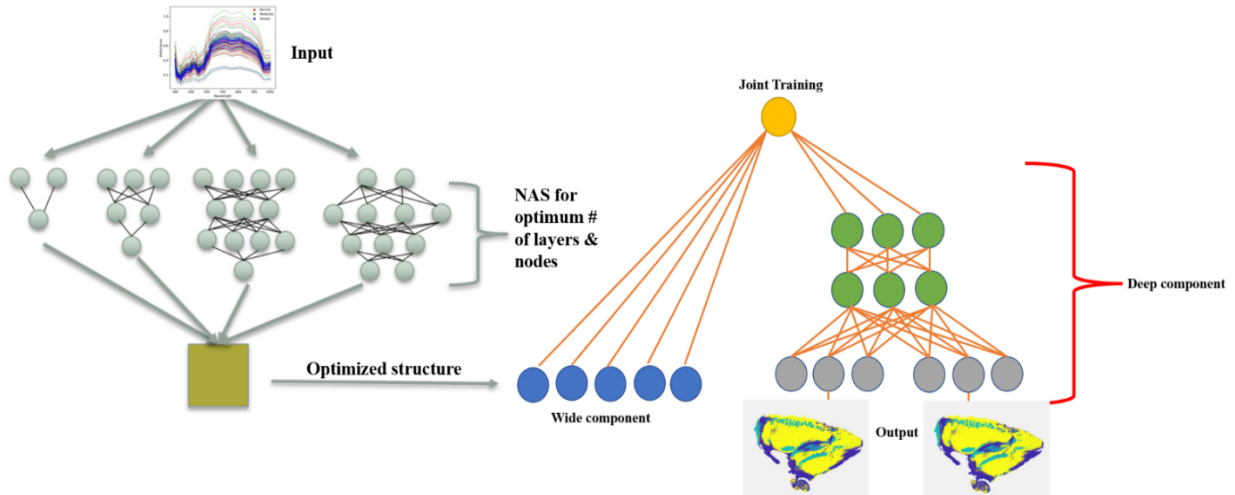

**Fig.1**. Neural network architecture search enabled wide & deep network model.

**2. Materials and methods:**

*2.1. Sample selection and data collection*

This study was conducted at the University of Arkansas's Biological and Agricultural Engineering laboratory. Two rounds of the experiment (round one: Nov 2022 and round two: June 2024) were conducted with a total of 125 broilers of 8 weeks of age randomly selected from the University's poultry farm and processed in the University of Arkansas pilot processing plant. Broilers were processed using commercial methods, and fillets were deboned from carcasses approximately 30 minutes after postmortem. Then, unchilled 250 chicken breast samples, including the left and right sides, were collected and scored in the processing plant. These samples were classified as the normal breast (NB), mild woody breast (MWB), and severe woody breast (SWB) using hand palpation by a specialist (NB=78; MWB=82; SWB=90; Fig.2). After WB scoring, these samples were moved to the laboratory for further hyperspectral imaging data collection. This processing protocol followed the procedure based on the previous study (Caldas-Cueva et al., 2021).

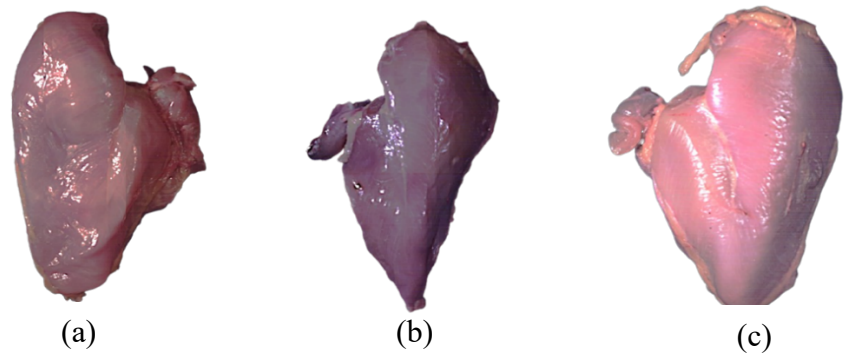

(a)    (b)    (c)

**Fig.2**. Raw breast fillets of (a) normal breast, (b) mild woody breast, and (c) severe woody breast.

The hyperspectral datasets were collected using the Specim LabScanner system (SPECIM FX10e, Oulu, Finland). This system is composed of a hyperspectral camera mounted 10 inches in height that captures wavelengths ranging from 397 to 1005 nm and a conveyor scanner with a 400×200 mm sample tray. The sample was put on a black plastic dish, and then the dish was placed on the tray for HSI scanning, as shown in Fig. 3. The illumination system includes dual halogen illumination (DECOSTAR 51 ALU 20W 12V 36deg GU5.3 halogen) with a lighting angle of about 45 degrees. This study followed the guidelines provided by Specim (Specim, 2020). Details about imaging pre-processing and calibration were elaborated in Section 2.2.

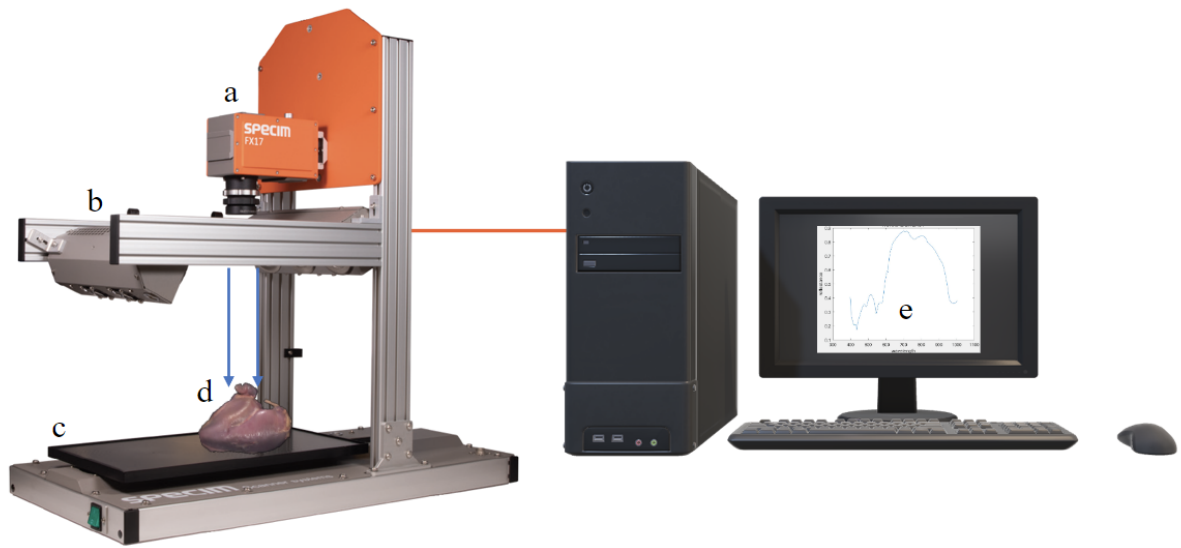

**Fig.3.** Hyperspectral imaging system setups: a) SPECIM FX 10e camera, b) Halogen illumination, c) Conveyor, d) chicken breast Sample, and e) spectral reflectance graph.

After the hyperspectral imaging scanning, hardness data of raw woody breast fillets were measured as the compression force (Newtons) per 1cm depth using a texture analyzer (MultiTest-dV-motorized texture analyzer, VectorPro software 6.16.0.0, Mecmesin, Slinfold, United Kingdom**).** The measurements used a 0.5cm flat probe to compress the fillet surface to 1 cm depth at the designated point without puncturing the tissue. Each fillet is divided into three regions:

cranial, medial, and caudal. Three points on each portion were measured, and 9 points were measured from each breast fillet (Fig. 4). The mean spectral signal of the 3 points for each region represents the corresponding region for further analysis.

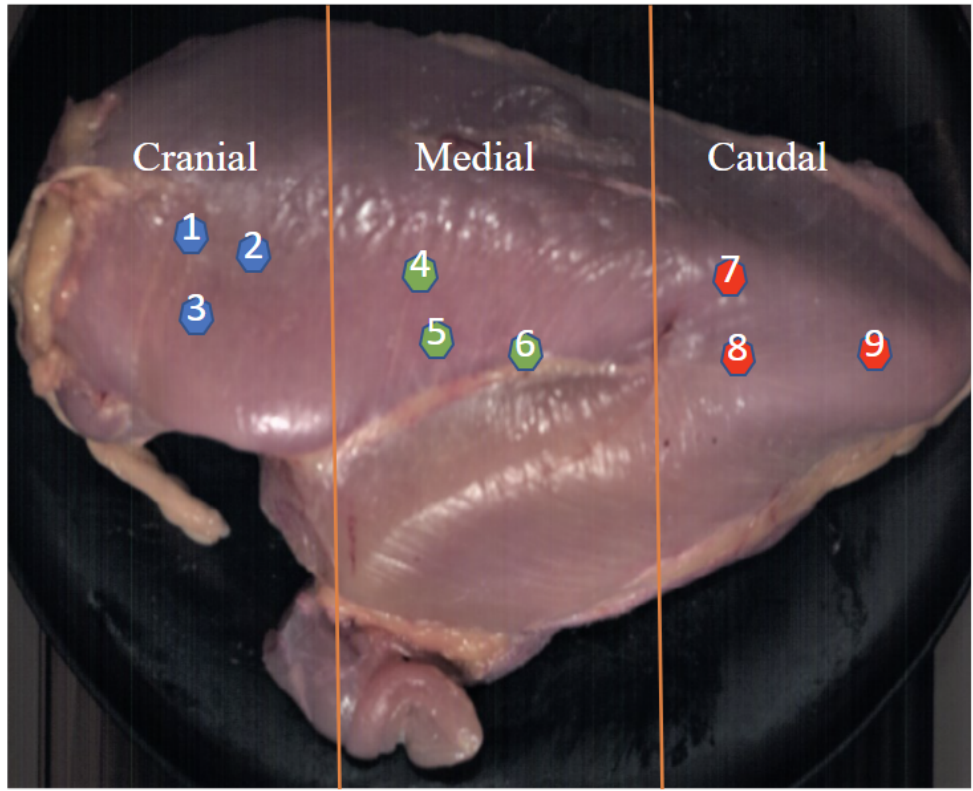

**Fig.4.** Raw breast fillet hardness analysis in three spatial regions (cranial, medial, and caudal). In three regions, 3 random number points are measured to represent the compression force value in the corresponding region.

## *2.2. Hyperspectral imaging acquisition and processing*

At the HSI data acquisition stage, LUMO scanner software (Specim Ltd, Finland) is used to adjust camera parameters and control the conveyor movement for image scanning. Hyperspectral data collection and processing involve a series of steps, including capturing and storing the raw hypercube, black-and-white calibration, spectral and spatial data pre-processing, and finally, spectral classification and regression (Fig. 5). In the experiment, the detailed motor parameters included acceleration and deceleration of 500 steps per second squared, a pitch of 10.24 millimeters, and positioning of one step. The camera had a spatial binning of 1, an exposure time of 10 seconds, and a frame rate of 50 fps.

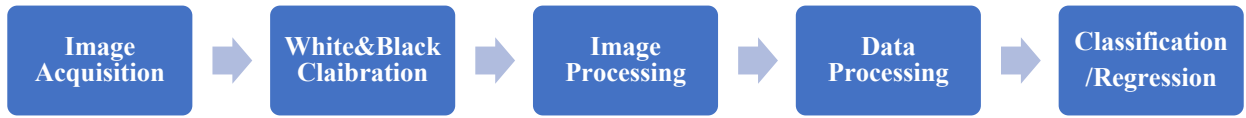

**Fig.5.** HSI collection and processing involved in woody breast analysis.

In the HSI pre-processing stage, the raw fillet hyperspectral image is pre-processed to extract refined reflectance data, which is used for classification and regression analysis. HSI pre-processing is a very important step to remove the noise and extract useful data (Elmasry et al., 2012), and it aims to standardize the spectral axis, validate the accuracy of spectral data, and minimize errors. In general, calibration involves both spectral waveband calibration and reflectance calibration. Each spectral channel is given a defined wavelength using a reference light source in spectral waveband calibration. Absolute standard reference spectra are used to calibrate the HSI system, which involves mapping obtained pixel intensities with their corresponding wavelengths using polynomial regressions. After capturing hyperspectral images of actual materials, the camera's background spectral response and dark current are taken into consideration through reflection calibration. The following formula is used to determine the pixel-based relative reflectance for the raw image using these reference images:

$$I = \frac{I' - D}{W - D} \qquad (i)$$

where D is the dark reference image, W is white reference image, I′ is the raw reflectance image, and I is the relative reflectance of images.

To extract the mean spectral of the fillet sample region, as shown in Fig. 6, a pseudo-RGB image is used and converted to L*, a*, and b* color space to generate a binary mask of breast fillet via the color thresholding method. The pre-processing step is implemented in both MATLAB software R2022b and Python 3.8.

Fig 7(a) shows the spectrum for each individual sample, which is used as an input for both WB level classification and texture hardness to generate the hardness map, as shown in Fig. 6. Fig 7(b) shows the mean spectrum in each WB group. From Fig 7(b), we found that the normalized spectral intensity in the severe WB (red) was consistently higher across most wavelengths, indicating distinct spectral characteristics likely due to significant structural or compositional changes. The mild WB breast (green) has moderate intensity between the other two types, but more fluctuations suggest milder changes in tissue properties. However, since spectral characteristics can vary from fillet to fillet, visual inspection alone is insufficient for accurate classification. Therefore, the proposed techniques are necessary for further analysis to accurately identify and categorize the fillets based on their spectral properties, ensuring robust and reliable detection of woody breast myopathy.

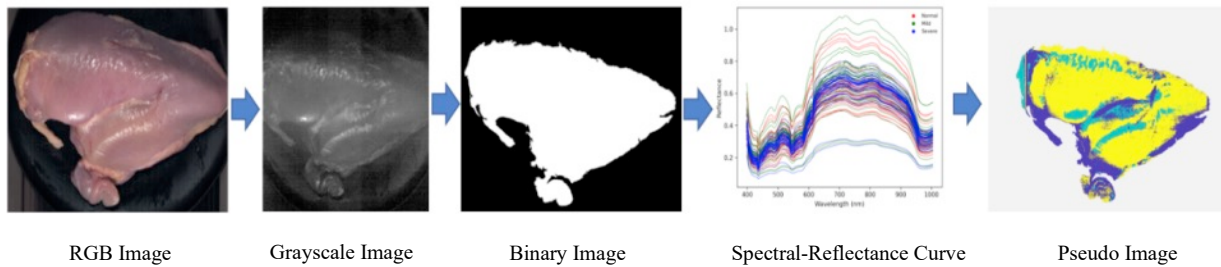

**Fig. 6.** Image processing steps and classification of the woody breast using hyperspectral imaging technology.

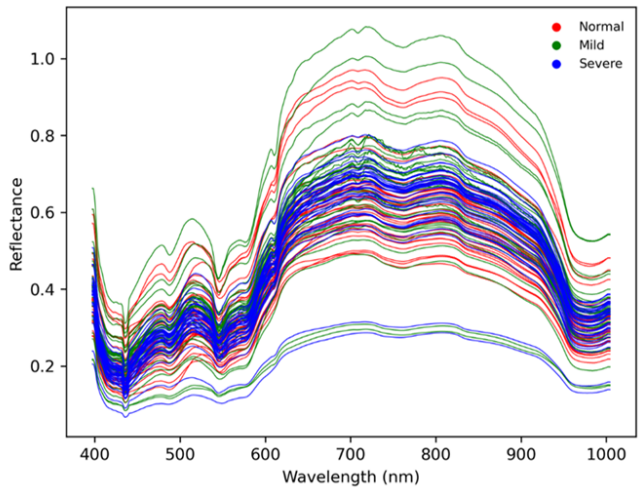
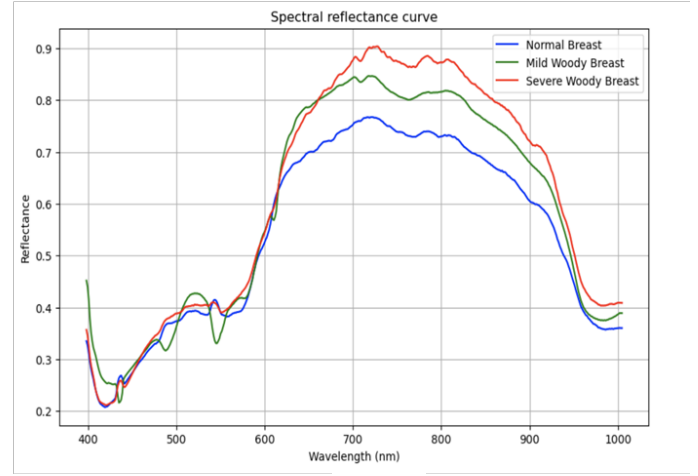

**Fig. 7.** (a) The raw spectral reflectance graph of chicken samples. (b) the average spectral reflectance graph in each WB group.

## *2.3. Neural network architecture search enabled wide deep learning for spectral signal classification and regression.*

To analyze the hyperspectral data, there is a need to build a mathematical model to describe the relationship between the input spectral signal and the outputs. In this study, two tasks need to be accomplished: woody breast level classification and textural compression force value regression. The classification task categorizes the chicken breast into NB, MWB, and SWB. The models took the spectral data, which means reflectance data in a specific region, as predictors and class labels or textural compression force as responses. After the model is trained, the model can be applied to other unknown samples for classification and regression purposes. In this research, NAS-WD is newly proposed to achieve the classification and regression task, the performance of which was compared with classic hyperspectral classification and regression models.

*2.3.1. Model descriptions*

The core idea of NAS-WD is based on the Multi-layer Perceptron model (MLP). MLP is a type of artificial neural network model that can be optimized, given or fine-tuned by adjusting the hyperparameters of the network to improve its performance on a particular task (Grossi and Buscema, 2007). However, there are two drawbacks to the generic MLP model. The first is that stacking multiple neural layers may diminish the capabilities of the model to capture the simple linear relationship between inputs and outputs, which may further lead to overfitting issues. The second drawback is that the network architecture will significantly affect the performance of the network. The designed NAS-WD model targets to solve the above two challenges by introducing wide-deep learning and neural network architecture search techniques.

*2.3.2. Wide and Deep learning*

The wide and deep learning techniques introduce two components in the network design, wide and deep components, to capture both shallow and deep relationships between the inputs and outputs. The wide component of the model consists of a basic linear model with the form $y = w^T x + b$, where 'y' represents the prediction, 'x' is a d-dimensional vector of features, 'w' is the learnable model parameter, and 'b' is the learnable bias term. In our experiment, the features used for the wide component include the original spectral features as the input.

The deep component is a feed-forward multi-layer neural network. Categorically related spectral features are converted into low-dimensional and dense feature vectors, also known as embedding vectors. These vectors are randomly initialized, fed into hidden neural network layers in the forward pass, and then trained to minimize the loss function. The prediction of the model is obtained by combining the output log odds of the wide and deep components using a trainable weighted sum to integrate both shallow and deep input features. This combined prediction is then passed through a logistic loss function for joint training. Joint training is a key feature of wide & deep learning which is different from ensemble training. In ensemble training, models are individually trained without knowing each other, which requires a large model size, whereas joint training allows for a smaller, wider part of the model to complement the deep part. Joint training is done by back-propagating the gradient output to the wide and deep parts of the model using stochastic optimization methods.

Wide and Deep Learning combines the strengths of both linear models (wide) and deep neural networks (deep). This architecture is particularly effective for tasks that involve both memorization and generalization, such as recommendation systems and ad click-through rate prediction. Fig. 8 is a detailed explanation of the workflow.

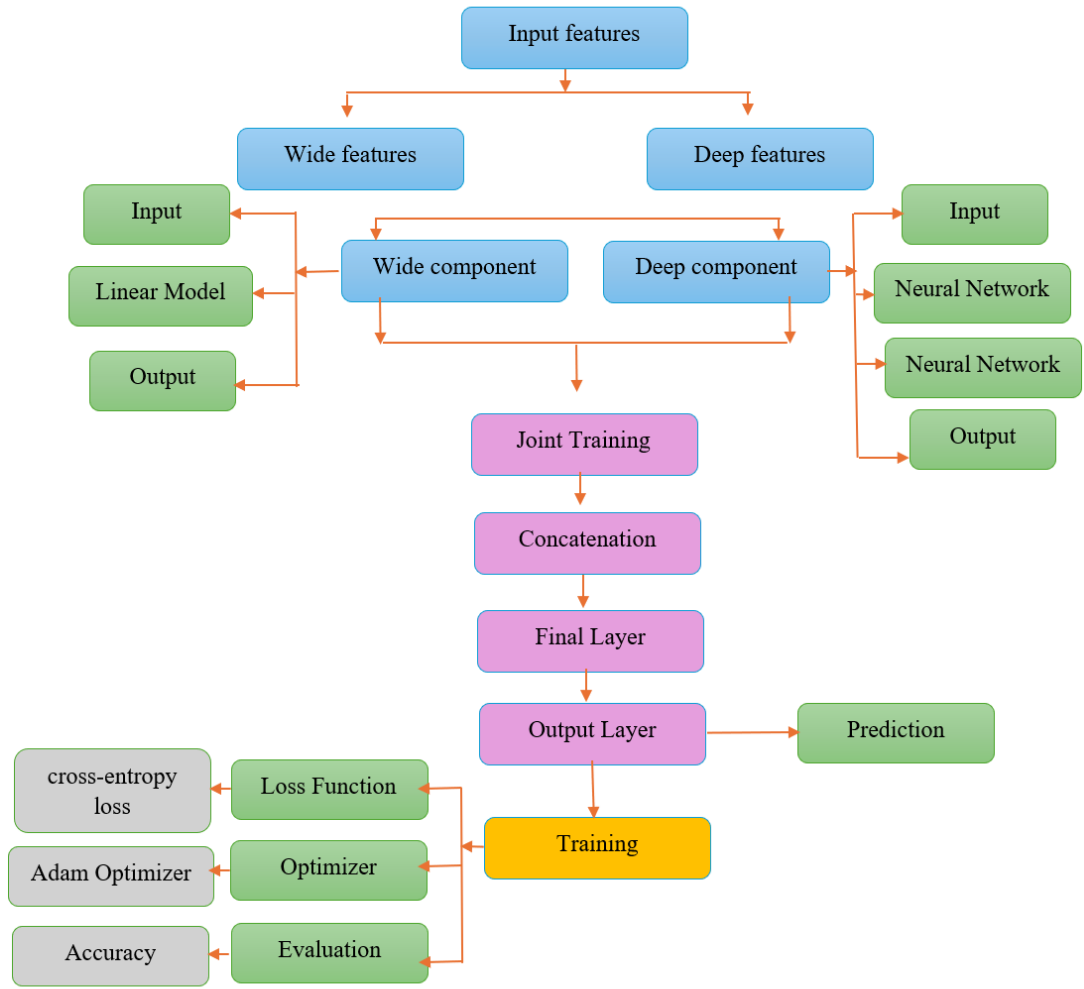

**Fig.8.** Workflow of Wide and Deep Learning.

The combination of memorization and generalization in WD learning (WDL) models leverages the wide component to capture feature interactions and memorize patterns directly from input features, while the deep component generalizes from data by learning high-level abstractions, allowing the model to handle both memorization (like a linear model) and generalization (like a deep neural network), which is beneficial for tasks with complex and diverse data patterns (Cheng et al., 2016). The model's end-to-end training, where both wide and deep components are optimized simultaneously using a single loss function, allows for better integration and coordination between components, leading to improved performance compared to separately trained models (Zhang et al., 2016). These highly flexible and customizable models allow for additional feature transformations, embeddings, and interaction terms, which better model complex relationships and interactions in data that traditional models may not capture effectively (Lian et al., 2018).

Designing and tuning WDL models can be complex due to the need to balance wide and deep components, requiring significant expertise and computational resources (Cheng et al., 2016). This complexity is managed by leveraging Neural Network Architecture Search (NAS), which automates the optimization of network architecture and hyperparameters, reducing development time and manual tuning efforts. Training WDL models can be computationally intensive, especially for large datasets with high-dimensional features, necessitating substantial computational resources (Guo et al., 2017). The computational intensity is addressed by joint training of both components, optimizing them simultaneously for efficient training. Hyperparameter sensitivity is mitigated by employing Bayesian Optimization (BO), which automates the search for optimal settings, reducing the need for extensive experimentation (Zhang et al., 2016). Wide and Deep Learning models are often seen as black boxes, which hinders interpretability in applications where understanding the decision process is crucial (Lipton, 2017). Regularization techniques like dropout, early stopping, and cross-validation are employed to prevent overfitting (Srivastava et al., 2014). While WDL models require large datasets, NAS-WD's ability to learn both linear and non-linear relationships makes efficient use of available data, enhancing performance even with limited data (Goodfellow et al., 2016). Despite longer training periods, NAS-WD models' advantages often compensate for this due to their automated hyperparameter tuning and ability to handle complex, high-dimensional data. While SVMs converge faster due to simpler optimization, they often fail to capture intricate relationships. MLPs, offering a middle ground, are more flexible than SVMs but typically don't match NAS-WD performance on complex tasks. In sum, NAS-WD models usually offer high predictive accuracy and can handle complex interactions, making them suitable for high-performance applications despite longer training times.

*2.3.3. Neural network architecture search*

Neural Architecture Search (NAS), whose general diagram is shown in Fig.9, has emerged as a powerful tool for automating the design of neural network architectures, exploring the space of possible network configurations to optimize performance for specific tasks. These algorithms automatically discover optimal architectures by tuning hyperparameters that describe the complex relations between variables, significantly outperforming manually designed models. For instance, (Zoph and Le, 2017) introduced a reinforcement learning-based NAS method that automates the design process, achieving state-of-the-art performance on several benchmarks. Additionally, (Liu et al., 2018) proposed a progressive approach to NAS, which reduces computational cost by gradually increasing model complexity, thereby making the search process more efficient.

Moreover, NAS methods can incorporate prior knowledge or leverage previous runs to accelerate convergence. This is particularly useful, as previous evaluations can inform future searches, reducing the overall computational expense. For example, (Real et al., 2019) demonstrated the effectiveness of using evolutionary algorithms in NAS, where the performance of previous architectures guides the search process. Similarly, (Pham et al., 2018) introduced an efficient NAS method using parameter sharing, significantly reducing search time by reusing parameters across different architectures during the search process. By leveraging existing knowledge and incorporating previous evaluations, NAS methods can more quickly and efficiently converge on optimal architectures, making them invaluable in the quest for advanced neural network designs.

In our NAS-WD network, the wide component employs a single dense layer, while the deep component uses multiple stacked dense layers. The NAS algorithms tune critical hyperparameters such as neuron node numbers, activation functions, the number of layers, and the dropout rate in the deep component. These hyperparameters are essential for accurately describing the complex relationships between variables, ultimately enhancing the model's performance.

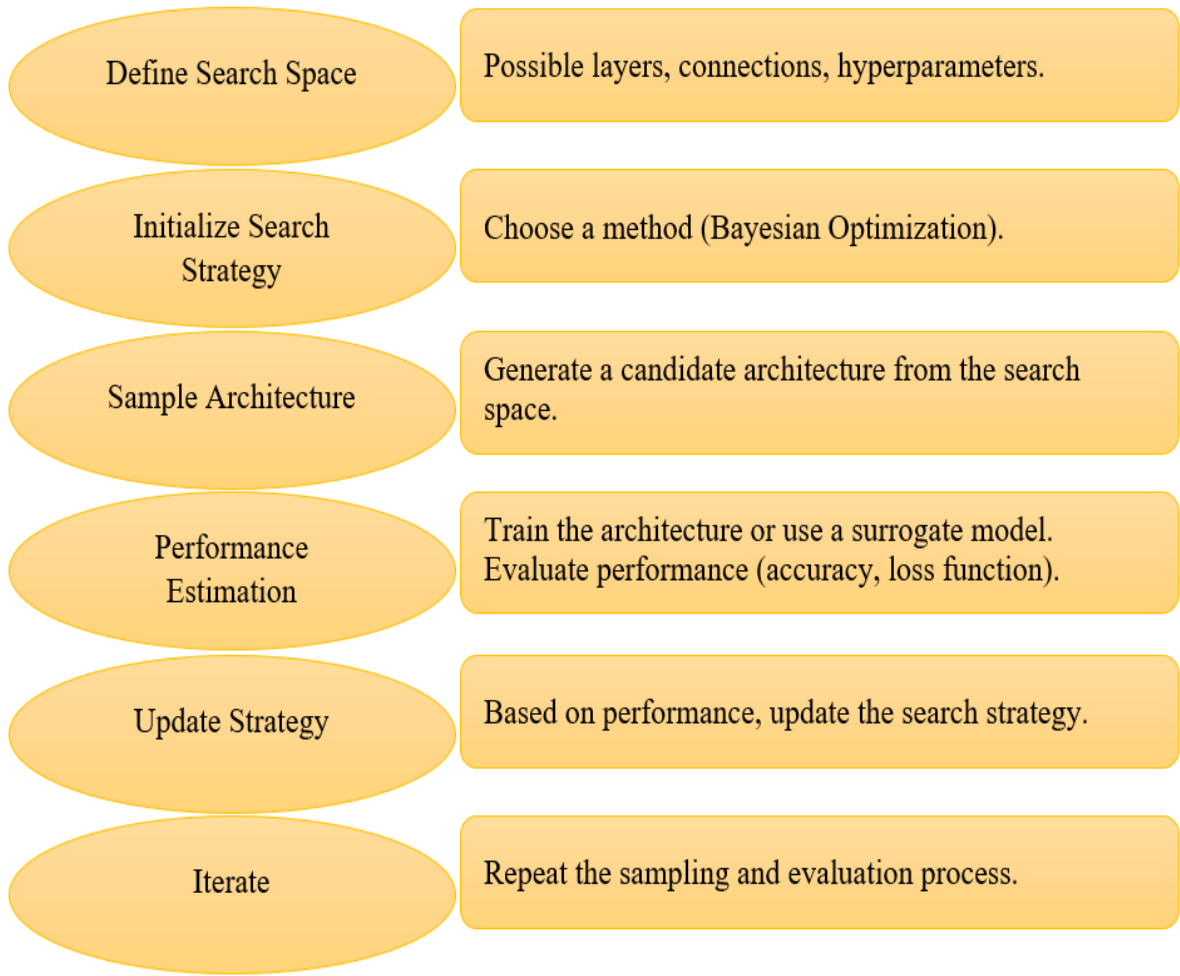

**Fig.9.** General diagram of NAS workflow

In addition, the outputs of both the wide and deep components are concatenated and the Adam optimizer was used as the network optimizer for the network training to predict the classes of woody breast myopathy. The learning rate of the Adam optimizer is also regarded as another hyperparameter that can be turned into the NAS model. The model is trained with early stopping to prevent overfitting, and the 5-fold cross-validation is performed to ensure the model's robustness across different data subsets. The choice of hyperparameters significantly impacts model performance, with activation functions affecting non-linear relationship capture, neural numbers controlling model capacity, dropout rate preventing overfitting, and learning rate influencing optimization speed. By tuning these hyperparameters, the best configuration for the dataset is identified, balancing complexity and generalization. Specifically, we use a Bayesian optimization tuner, one of the NAS models, to efficiently search the search space to find the best network architecture and optimize the learning rate to get the top-performing model.

*2.3.4. Bayesian Optimization*

Bayesian Optimization is an efficient strategy for optimizing expensive black-box functions (Turner et al., 2021). It's widely used for hyperparameter tuning in machine learning, including Neural Architecture Search (NAS). Here's a detailed explanation of the workflow (Fig. 10).

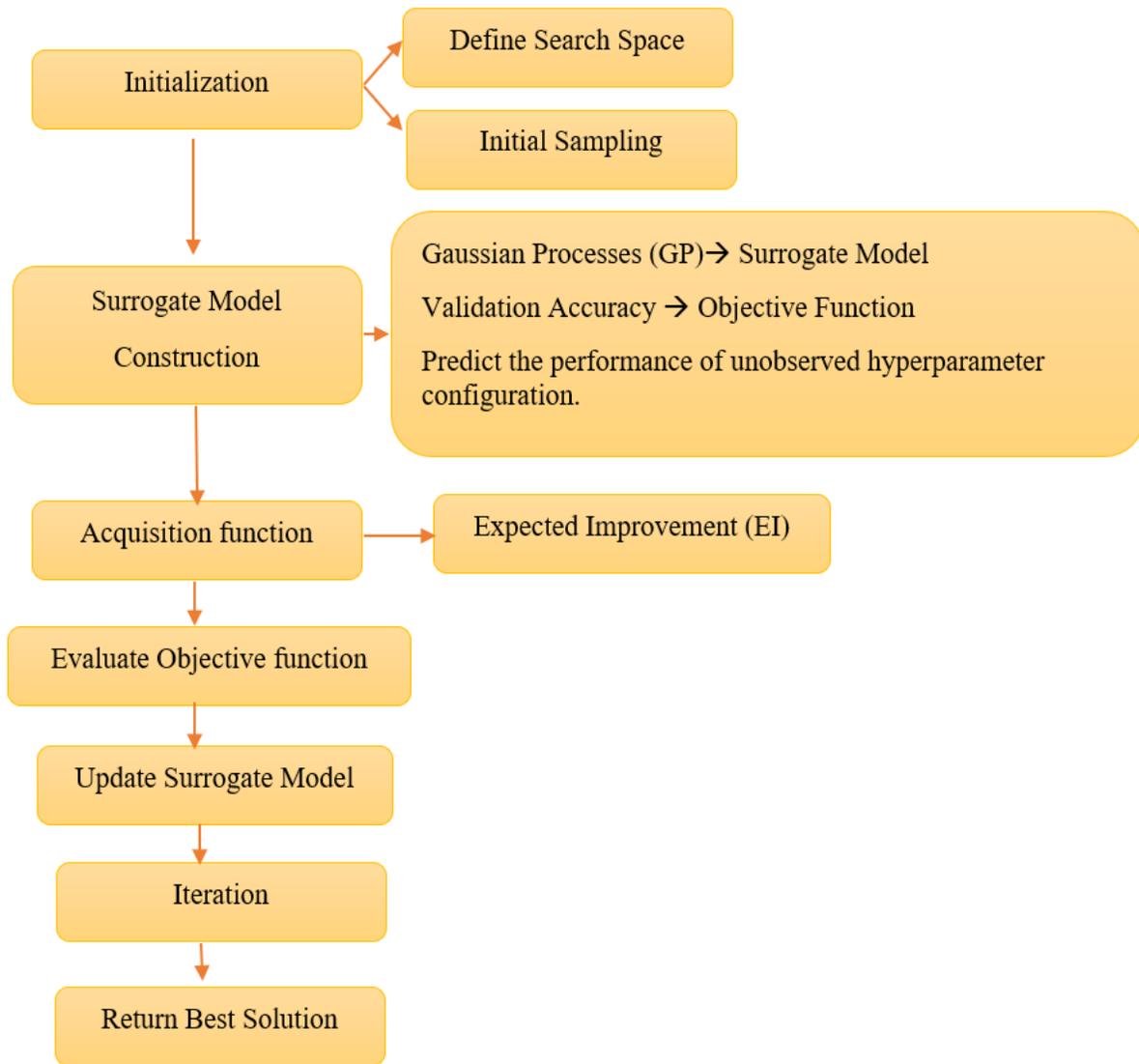

**Fig.10.** Workflow of Bayesian Optimization.

Bayesian Optimization utilizes a surrogate model, typically Gaussian Processes (GP), to approximate the objective function. Gaussian Processes (Williams and Rasmussen, 2006) are commonly used because they provide uncertainty estimates for predictions. The objective function in this Bayesian Optimization process is validation accuracy, which assesses the performance of different hyperparameter configurations by measuring accuracy on the validation set. The process involves fitting a Gaussian Process to the observed data, using the Gaussian Processes to make predictions for new points, and employing an acquisition function to select the next point to evaluate (Turner et al., 2021). This approach allows for efficient exploration and exploitation of the search space to find the optimum of the objective function.

Bayesian Optimization is highly advantageous over other search strategies due to its sample efficiency and reduced number of evaluations, which lower computational cost (Li et al. 2018) and time compared to random or grid search (Bergstra et al. 2011)). Creating a surrogate model to approximate performance predicts new configurations' outcomes without exhaustive evaluations, focusing on promising regions. Unlike local search methods, Bayesian Optimization aims for global optimization using an acquisition function to navigate complex, high-dimensional spaces. It accommodates both discrete and continuous hyperparameters, making it versatile for various optimization problems. Compared to Reinforcement Learning (RL), Bayesian Optimization is less resource-intensive and simpler to implement, avoiding the high computational demands of RL (Snoek et al. 2012). It achieves efficiency faster than Evolutionary Algorithms, which often require many generations to converge (Real et al. 2019).

*2.3.5. Implementation details*

The NAS-WD model, which comprises a wide component (single dense layer) and a deep component (multiple stacked dense layers), was optimized using Keras Tuner with Bayesian Optimization. Python 3.8 and TensorFlow 2.4.1 were used in network implementation. The search space for NAS included hyperparameters such as activation functions (ReLU, sigmoid), number of units (32 to 512), number of layers (1 to 3), dropout rates (0.0 to 0.5), and learning rates (1e-4 to 1e-2). The model parameters involved defined both the wide and deep branches, and the Adam optimizer. The model was compiled with Sparse Categorical Cross-entropy loss as the evaluation metric for the classification task and mean squared loss as the evaluation metric for the regression task. The training process used early stopping with the patience of 10000 epochs, and cross-validation was performed using 5-fold cross-validation. During cross-validation, the model was trained and validated on splits of the training data with a fixed random seed for reproducibility and performance comparison. The performance metrics such as accuracy and confusion matrix were recorded for each fold, whose average will be reported in the next section.

.

## 3. Results

### *3.1. Sample hardness distribution*

Hardness is a very important characteristic in woody breast myopathy, which can be used to classify samples into different severity levels it belongs to. Hardness is heterogeneous throughout the chicken breast fillet surface, making automatic and instrumental-based one-point-based WB classification challenging (Wold et al., 2019). In the experiment, each sample was divided into cranial, medial, and caudal portions. The texture analyzer collected the compression force values for NB, MWB, SWB samples region by region. As shown in Fig. 2, three random points are collected in each region, and the average compression force value of the three points represents the compression force value of that region. Fig 11 shows the average compression force value of the total nine points representing the hardness value of the whole fillet. ANOVA test shows that the average compression force values for the three categories differed significantly ($p < 0.05$). Fig.12 shows the average compression force values for the three categories at three different portions. If the ANOVA test was conducted region by region (cranial to cranial, medial to medial, caudal to caudal) and the results showed that only the cranial portion of the samples was significantly different ($p < 0.05$) compared to medial and caudal. The mean and standard deviation of compression force values of the whole breast fillet and each portion were presented in the Table 1, showing that the mean values were spatially heterogeneous. In the following harness regression analysis, only the cranial region's compression force values were considered.

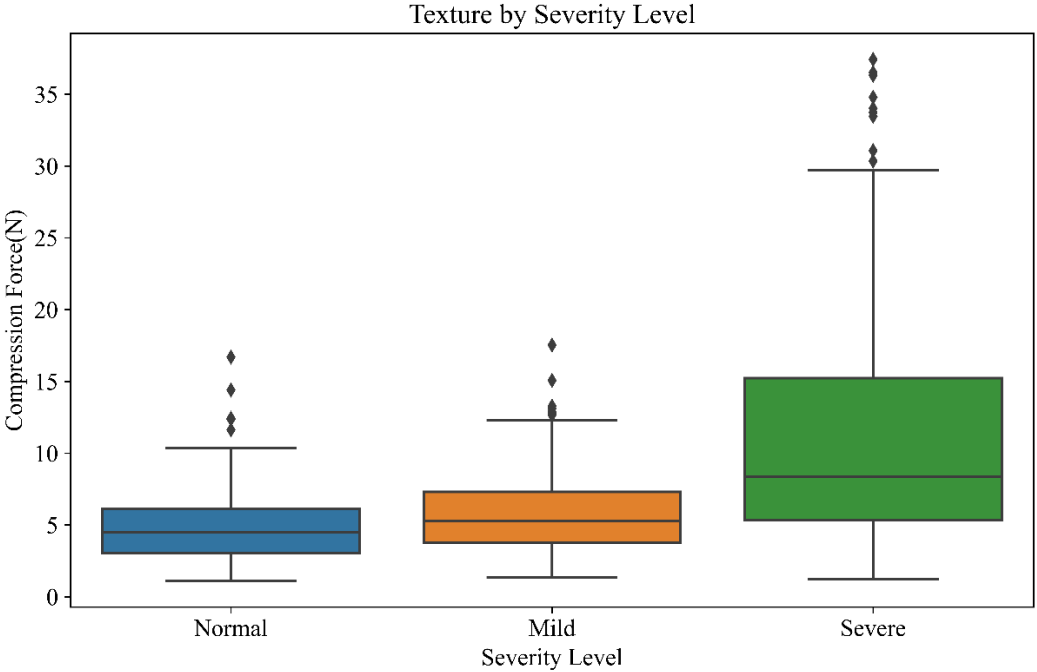

**Fig.11.** Compression force of raw whole woody breast fillets in different woody breast severity groups.

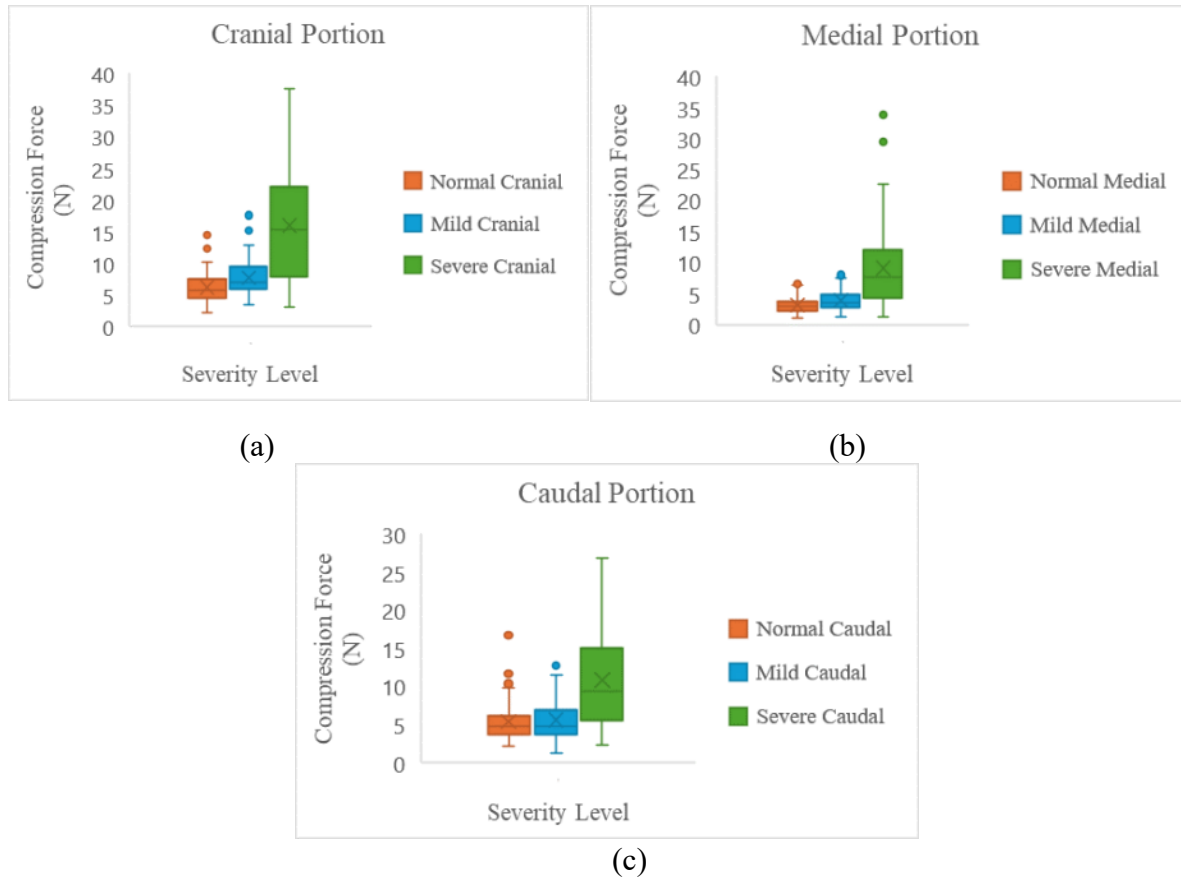

**Fig. 12.** Compression force of raw fillets of different WB levels in (a) cranial portion, (b) medial portion, and (c) caudal portion.

**Table 1.** Compression force values of raw whole woody breast fillets in Newtons (N).

| Severity level | Whole | Cranial | Medial | Caudal |
| --- | --- | --- | --- | --- |
| Normal Breast | 4.89 ± 2.47 | 7.02 ± 2.23 | 3.14 ± 1.21 | 5.33 ± 2.70 |
| Mild Woody Breast | 5.47 ± 2.78 | 8.23 ± 2.50 | 4.06 ± 1.54 | 4.78 ± 3.02 |
| Severe Woody Breast | 11.12 ± 7.21 | 21.03 ± 6.75 | 11.16 ± 6.42 | 13.59 ± 6.26 |

*3.2. WB severity classification analysis*

For the classification task, we have evaluated NAS-WD, SVM, and MLP in our dataset. A fixed state parameter ensures reproducibility in the experiment by leading to consistent dataset splits each time the code runs. In the experiment, 5-fold cross-validation was conducted, meaning the dataset was divided into five equal parts, and the model training and validation process ran five times. Each time uses a different subset as the validation set and the remaining four for training. This approach ensures that each fold is evaluated once, providing a comprehensive and balanced model evaluation. The classification results of NAS-WD were compared with the SVM and MLP results, which showed NAS-WD model can achieve higher training and test accuracy (Table 2).

**Table 2.** Performance comparison of different machine learning models for the WB level classification task[#].

| Model | Training Accuracy | Test Accuracy | Confidence Interval | Test Precision | Test Recall | Test F1 Score |
|---|---|---|---|---|---|---|
| NAS-WD* | **97.50** | **95.00** | **(0.89, 1.00)** | **95.22** | **95.00** | **0.9496** |
| SVM | 94.90 | 80.00 | (0.77, 0.82) | 80.27 | 80.00 | 0.7945 |
| MLP | 78.10 | 72.80 | (0.61, 0.84) | 73.48 | 72.80 | 0.7219 |

[#] All reported metrics are the average of the five-fold cross-validation; * NAS-WD: Neural Network Architecture Search Enabled Wide-Deep Learning Model

The NAS-WD model demonstrates superior performance with the highest training accuracy (97.50%) and test accuracy (95.00%), indicating excellent training and generalization capabilities. Its test Precision (95.22%), Recall (95.00%), and F1 Score (0.9496) are also the highest among the three models, showcasing outstanding accuracy and balance. The Confidence Interval (CI) for NAS-WD is 0.89 to 1.00, reflecting high reliability in its accuracy. In contrast, while having a decent training accuracy (94.90%), the SVM model shows a lower test accuracy (80.00%) with a CI range of 0.77 to 0.82, indicating potential overfitting and lower generalization. Its test Precision (80.27%), Recall (80.00%), and F1 Score (0.7945) are also lower than those of NAS-WD. In the experiment, grid search algorithms are applied to the SVM model to identify the best SVM kernels and hyperparameters, and the best performances are reported above. The MLP model performs the worst, with the lowest training accuracy (78.10%) and test Accuracy (72.80%), suggesting it struggles with both training and generalization. Its test Precision (73.48%), Recall (72.80%), and F1 Score (0.7219) are the lowest among the three models, and its CI range of 0.61 to 0.84 indicates higher variability. The results from MLP show the importance of NAS design. Overall, NAS-WD is the best model due to its highest performance metrics and narrow CI range, demonstrating robust performance and excellent generalization. The confusion matrix of these models on test data is shown in Fig. 13.

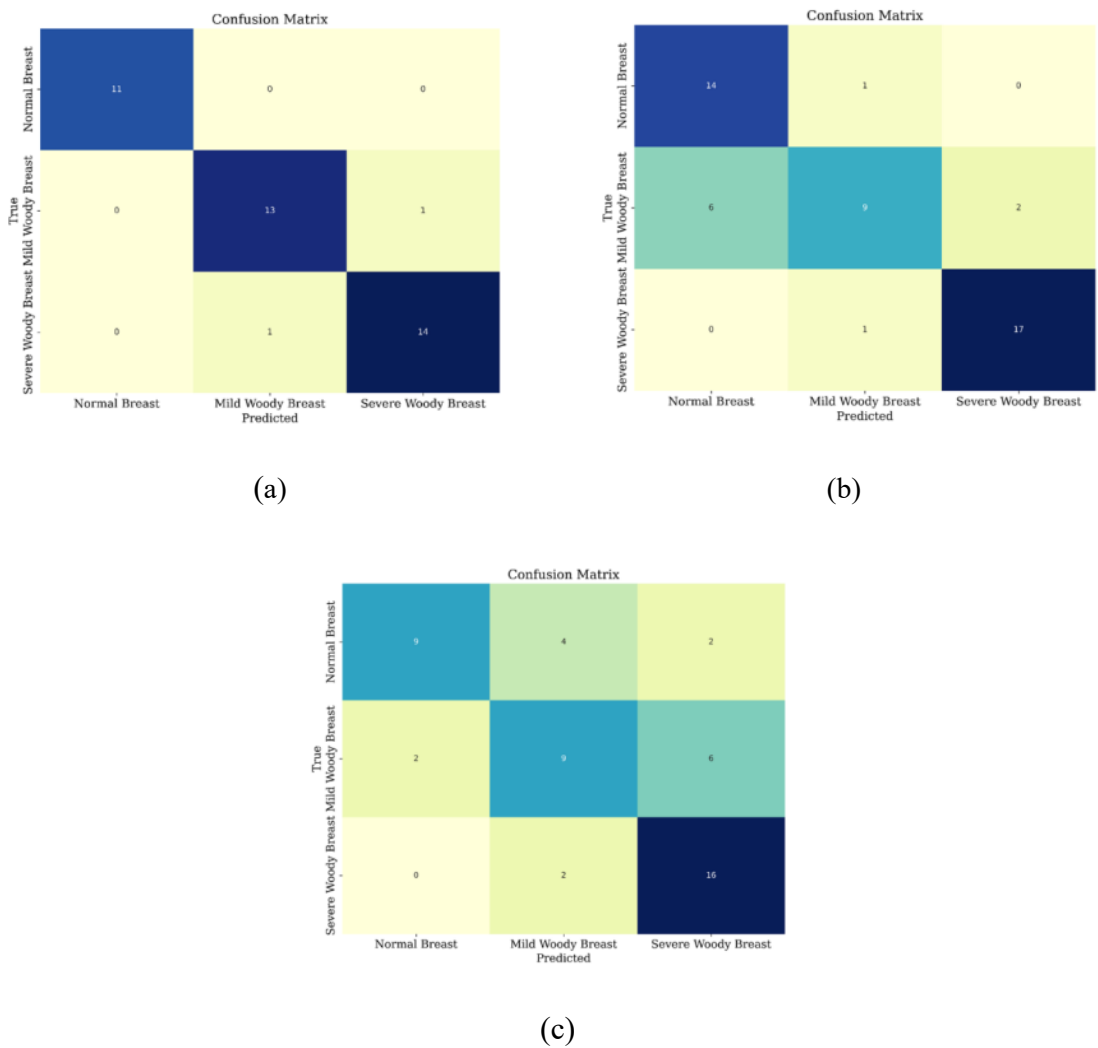

**Fig.13.** Confusion matrices for the WB classification task across three models: (a) NAS-WD, (b) SVM, and (c) MLP, based on 50 test samples in one cross-validation fold.

With the well-trained NAS-WD, we applied the model to predict categories of each pixel in a single HSI cube to visualize the prediction distribution of each sample, as shown in Fig 14. Fig. 14(a1), 14(b1), and 14(c1) are the classified pseudo images in which the severity level is represented in three colors: green for NB, yellow for MWB, and red for SWB indicating severity scales (0-2) respectively and simultaneously their respective pie charts are shown in Fig. 14 (a2), (b2), and (c2) indicating the percentage of pixels that were classified as NB, MWB, and SWB.

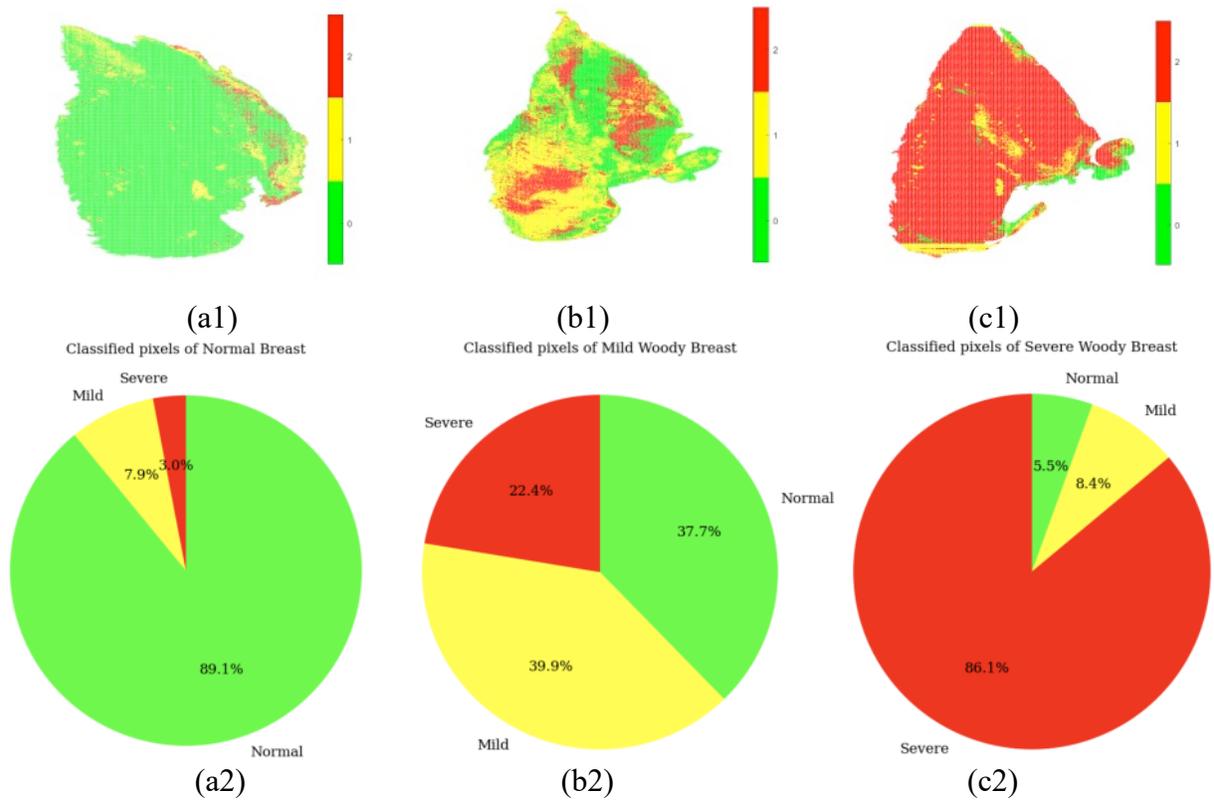

**Fig.14.** Classified pseudo images of NB (a1), MWB (b1), and SWB (c1) with respective pie charts (a2, b2, c2) pixel classification percentages. The classification pie charts are the indicators of the percentage of the severity level of the woody breast condition we can be used to decide whether the woody breast condition is normal, mild, or severe.

*3.3. Fillet hardness regression analysis*

In the regression test, the correlation between hyperspectral signals and compression force values was first established via the NAS-WD model. In the study, the compression force values above 10.8 N were considered outliers and removed for analysis based on previous studies related to instrumental texture characteristics and instrumental compression force (Chatterjee et al., 2016; Sun et al., 2018). As mentioned in Section 3.1, because the compression force values in the cranial portion values were significantly different across different WB categories compared to the other two regions, we chose to correlate only the cranial portion of hardness with the spectral data. The cranial region can be easily segmented out from an image processing algorithm. As the comparison, the performance of NAS-WD is compared with SVR and PLSR model, as shown in Fig. 3. The SVR model (Table 3), which uses hyperparameter grid search to identify the best model (rbf kernel, gamma: 1, C: 10), and same for the PLSR (10 principal components).

**Table 3**. Performance comparison of different machine learning models for the fillet hardness regression task.

| Model | Correlation Coefficient (r) | $R^2$ | RMSE |
|---|---|---|---|
| NAS-WD | 0.75 | 0.514 | 0.521 |
| SVR | 0.56 | 0.15 | 0.678 |
| PLSR | 0.51 | 0.208 | 0.920 |

Among the models evaluated, NAS-WD emerges as the best-performing model based on the provided metrics. It exhibits the highest correlation coefficient of 0.75 and $R^2$ value of 0.514, indicating a strong linear relationship between predicted and actual values, which is substantially higher compared to SVR and PLSR. In addition, the root mean squared error (RMSE) from NAS-WD is also lower compared to the other models, indicating a more accurate and reliable prediction overall. The well-trained NAS-WD model was then used to create a visual hardness distribution map based on the spatial data collected from the hyperspectral imaging. This hardness distribution maps shown in Fig 15 (a1), (b1), (c1) represent the predicted compression force values at different portions of the chicken breast fillet at the pixel level, demonstrating the practical application of the NAS-WD model in providing detailed insights into the texture variations of the meat. In addition, the pie chart visualization and hardness distribution map (Fig 15 (a2), (b2), (c2)) show the percentage of pixels that were predicted in the range of 0 to 3.5 N, 3.6 to 7.1 N, 7.11 to 10.8 N, and > 10.8 N. Based on the percentages obtained, we can also classify the woody breast fillets into three classes, but this may not be solely reliable yet as r = 0.75.

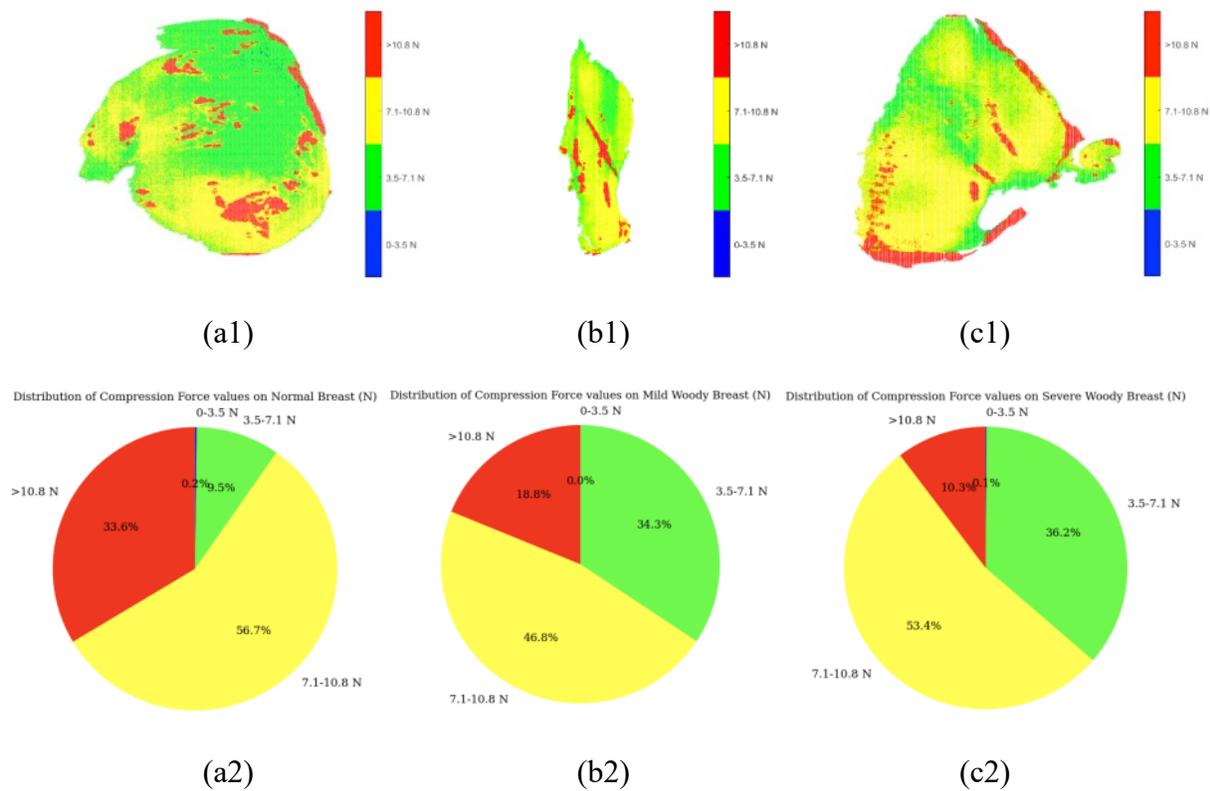

**Fig.15.** Hardness distribution maps of NB (a1), MWB (b1), and SWB (c1) with respective pie charts (a2, b2, c2) pixel classification percentages of different hardness levels.

## 4. Discussion

The findings from our study align with existing literature on the importance of hardness in characterizing WB myopathy and its influence on meat quality. Previous studies identified the hardness of WB fillets, highlighting that WB muscles exhibit a unique ridge and a contracted muscle appearance (Mudalal et al., 2015; Sihvo et al., 2014). Our results corroborate these findings by showing significant differences in compression force values between normal, moderate, and severe WB categories, particularly in the cranial portion of the fillet. Additionally, our data (Table 1) is consistent with prior studies, which indicate significant changes in hardness in specific regions of the WB: the cranial portion for mild WB and all regions (cranial, medial, and caudal) for severe WB (Chatterjee et al., 2016; Sun et al., 2018; Tijare et al., 2016).

However, our research offers several advancements over previous studies. Unlike prior work that primarily focused on single-point measurements, we calculated the compression force for the entire fillet by averaging values from multiple regions (cranial, medial, and caudal). This approach provides a more comprehensive assessment of the fillet's hardness distribution, capturing its heterogeneity more effectively. Additionally, our study is pioneering in correlating hyperspectral data with compression force values to generate a hardness distribution map, offering a novel WB classification and assessment method.

For the woody breast classification task, results showed superior performance compared to different previously reported methods beyond hyperspectral imaging. Siddique et al. (2022) used an SVM model with bioelectrical impedance analysis to classify WB severity, achieving accuracies of 77.78%, 85.71%, and 88.89% for normal, mild, and severe WB, respectively. However, the variability in accuracy across severity levels suggests potential inconsistencies in model performance. The reliance on hand-held devices introduces user-dependent variability and measurement inconsistencies, and the limited feature set may not capture the full complexity of WB compared to hyperspectral imaging (HSI). Yang et al. (2021) employed a deep learning model (AlexNet) with the expressible fluid method, achieving 93.3% accuracy for fresh and 92.3% for frozen fillets, indicating the model's performance can be affected by sample state, potentially limiting generalizability. Additionally, AlexNet's computational intensity requires significant resources for training and inference, and the method's accuracy is highly dependent on the specific expressible fluid technique, which may not be universally applicable. Yoon et al. (2022) focused on binary classification based on bending properties, achieving over 95% accuracy; however, this approach does not differentiate between multiple WB severity levels, limiting its applicability for detailed grading. Furthermore, specialized equipment for bending properties measurement may not be practical for large-scale applications, and the high accuracy might indicate potential overfitting to the specific dataset, reducing effectiveness on unseen data or in different settings. At the same time, using HSI data with the NAS-WD algorithm showed a classification accuracy of 95% for three classes in this study, which outperforms most existing studies.

From another aspect, the study of the relationship between the HSI data and hardness has never been reported in previous literature. This study first considered the heterogeneous behavior of hardness on the fillet's surface. Specifically, the various amounts of water in the tissues could induce changes in hardness measurement and the intensity of reflectance signals. Besides the water content, many other chemical components will affect the hyperspectral signals; therefore, the spectral information is not directly or linearly related to hardness, which might be the reason why the classic models SVR and PLSR were not able to establish such a complex relationship between the hardness and spectral data. While the NAS-WD model, having both linear and non-linear behaviors along with joint training, established a pretty good relationship between the spectral signals and hardness with a correlation coefficient of $r = 0.75$, demonstrated superior predictive power compared to traditional models like SVR and PLSR. In further studies, it would be worthwhile exploring the NAS-WD model with different search optimization techniques to increase the accuracy of detection of mild woody breast and correlating the HSI data with hardness data either by increasing the number of samples or by introducing another variable like water holding capacity, which could be the factor directly related to changes in reflectance data.

Even the developed NAS-WD model has achieved superior performance in WB severity classification and hardness regression tasks, manual featuring engineering has proven beneficial over automatic feature learning methods in various food engineering applications, especially the involvement of domain-specific knowledge allowed the model to distinguish subtle differences in known content that automatic features might miss for better interpretability (Mukhiddinov et al., 2022, Ali et al., 2020). Considering that the woody breast and the hardness of the fillet are not determined by a single factor, automatic feature learning has its advantages, but our future study will consider integrating the domain knowledge to improve the model's robustness.

## 5. Conclusion

This study uses HSI to introduce a novel NAS-enabled NAS-WD to classify woody breast conditions in chicken fillets. The NAS-WD model demonstrated superior accuracy (95%) and handling of complex non-linear relations compared to traditional models like SVM (80%) and MLPs (72.8%). It achieved a regression correlation of 0.75 with hardness data, outperforming SVR (0.56) and PLSR (0.51). Uniquely, it accounted for the spatial heterogeneity of fillet hardness and generated hardness distribution maps. Future research should explore the impact of chilling on HSI data and incorporate water loss into regression models. In conclusion, the NAS-WD model offers an efficient, accurate, and non-invasive method for detecting woody breast conditions, promising significant cost savings and improved quality in the poultry industry.

**Acknowledgement**

The authors acknowledge the grant support by the University of Arkansas Experimental Station and the University of Arkansas College of Engineering, USDA National Institute of Food and Agriculture (No: 2023-70442-39232, 2024-67022-42882).